\documentclass{article}

\usepackage[letterpaper]{geometry}

\usepackage{times}
\usepackage{soul}
\usepackage{url}
\usepackage[utf8]{inputenc}
\usepackage[small]{caption}
\usepackage{graphicx}
\usepackage{amsthm}
\usepackage{booktabs}
\usepackage{amssymb}
\usepackage{amsmath}
\usepackage{mathtools}
\usepackage{amsfonts}
\usepackage{bm} 
\usepackage{graphicx}
\usepackage{rotating}
\usepackage{tikz}
\usetikzlibrary{bayesnet}
\usepackage{tabularx,hhline}
\usepackage{array}
\usepackage{tabulary}
\newcolumntype{K}[1]{>{\centering\arraybackslash}p{#1}}
\usepackage{array}
\usepackage{multirow}
\usepackage{caption}
\usepackage{subcaption}
\usepackage{tabularx,booktabs} 
\usepackage{standalone}
\usepackage[acronym,smallcaps,nowarn]{glossaries}

\usepackage{times}  %Required
\usepackage{graphicx}  %Required
\usepackage{listings}
\usepackage{fancyvrb}
\fvset{fontsize=\normalsize}

\usepackage{hyperref}

\usepackage{algorithm,algorithmic}

\usepackage{enumerate}
\usepackage[inline]{enumitem}
\usepackage{url}

\usepackage{bbm}

\usepackage{ltexpprt}
\usepackage{hyperref}
\usepackage[numbers]{natbib}
\DeclareMathOperator*{\argmin}{\mathrm{argmin}}

\usepackage[multiple]{footmisc}
\usepackage{bigfoot}

\DeclareNewFootnote{AAffil}[arabic]
\DeclareNewFootnote{ANote}[fnsymbol]

\begin{document}

\newcommand\relatedversion{}
\renewcommand\relatedversion{\thanks{The full version of the paper can be accessed at \protect\url{https://arxiv.org/abs/1902.09310}}} % Replace URL with link to full paper or comment out this line

\title{\Large Estimating Latent Population Flows from Aggregated Data via Inversing Multi-Marginal Optimal Transport}
%\author{Sikun Yang\thanks{Shenzhen Institute of Artificial Intelligence and Robotics for Society; The Chinese University of Hong Kong, Shenzhen. Email: yangsikun@cuhk.edu.cn}
%\and Hongyuan Zha\thanks{Shenzhen Institute of Artificial Intelligence and Robotics for Society; The Chinese University of Hong Kong, Shenzhen. Email: zhahy@cuhk.edu.cn}}
%\author{Sikun Yang\thank{School of Data Science, Shenzhen Institute of Artificial Intelligence and Robotics for Society, The Chinese University of Hong Kong, Shenzhen. Email: \{yangsikun, zhahy\}@cuhk.edu.cn}
%\and Hongyuan Zha$^*$
%%\thanks{Shenzhen Institute of Artificial Intelligence and Robotics for Society; The Chinese University of Hong Kong, Shenzhen. Email: zhahy@cuhk.edu.cn}
%}
\author{ {\bf Sikun Yang}, {\bf Hongyuan Zha}\\
Shenzhen Institute of Artificial Intelligence and Robotics for Society\\
The Chinese University of Hong Kong, Shenzhen \\
\texttt{\{yangsikun, zhahy\}@cuhk.edu.cn}\\
%\And
%{\bf Heinz Koeppl}  \\
%% ETIT Department        \\
%Technische Universit\"at Darmstadt \\
%Darmstadt, Germany\\
%\texttt{heinz.koeppl@bcs.tu-darmstadt.de}
%\And
%{\bf Coauthor}   \\
%Affiliation \\
%Address    \\
%(if needed)\\
}

\date{}

\maketitle

% Copyright Statement
% When submitting your final paper to a SIAM proceedings, it is requested that you include
% the appropriate copyright in the footer of the paper.  The copyright added should be
% consistent with the copyright selected on the copyright form submitted with the paper.
% Please note that "20XX" should be changed to the year of the meeting.

% Default Copyright Statement
%\fancyfoot[R]{\scriptsize{Copyright \textcopyright\ 2023 by SIAM\\
%Unauthorized reproduction of this article is prohibited}}

% Depending on which copyright you agree to when you sign the copyright form, the copyright
% can be changed to one of the following after commenting out the default copyright statement
% above.

%\fancyfoot[R]{\scriptsize{Copyright \textcopyright\ 20XX\\
%Copyright for this paper is retained by authors}}

%\fancyfoot[R]{\scriptsize{Copyright \textcopyright\ 20XX\\
%Copyright retained by principal author's organization}}

%\pagenumbering{arabic}
%\setcounter{page}{1}%Leave this line commented out.

\begin{abstract}
We study the problem of estimating latent population flows from aggregated count data. This problem arises when individual trajectories are not available due to privacy issues or measurement fidelity. Instead, the aggregated observations are measured over discrete-time points, for estimating the population flows among states. Most related studies tackle the problems by learning the transition parameters of a time-homogeneous Markov process.  
Nonetheless, most real-world population flows can be influenced by various uncertainties such as traffic jam and weather conditions. 
Thus, in many cases, a time-homogeneous Markov model is a poor approximation of the much more complex population flows. To circumvent this difficulty, we resort to a multi-marginal optimal transport (MOT) formulation that can naturally represent aggregated observations with constrained marginals, and encode time-dependent transition matrices by the cost functions. In particular, we propose to estimate the transition flows from aggregated data by learning the cost functions of the MOT framework, which enables us to capture time-varying dynamic patterns. The experiments demonstrate the improved accuracy of the proposed algorithms  than the related methods in estimating several real-world transition flows. 
\end{abstract}

%\linespread{0.99}
%\input{section_introduction}
\section{Introduction}

This work focuses on the problems where 
data about individuals are not readily available because of various reasons such as privacy issues and measurement fidelity. Instead, we only have access to the population-level aggregate data that could be incomplete and noisy. For instance, when studying the infectious disease spreading~\cite{COVID}, it is too expensive or even impossible to track the trajectory of each individual. Nevertheless, the number of individuals in some regions over discrete time points, can be measured using sensing devices. Statistical analysis of these aggregate data is challenging, and has received amounts of attention in diverse fields including estimating ensemble flows~\cite{HaaslerRCK19,MSJ2021}, steering opinion dynamics among humans~\cite{WangTVS16}, epidemic forecasting~\cite{EF} among others.  

Over last decade, many efforts, such as collective graphical models (CGMs)~\cite{CGMs,ApproxCGMs,MPCGMs,MMC}, have been dedicated to the problem of inference and learning with aggregated data. These methods often assume that the individuals behind aggregated data, behave according to a time-homogeneous Markov chain. However, in many cases, the individual movement behaviors are significantly affected by various factors including weather conditions, traffic situations, and so on. Hence, estimating latent transition flows only with a time-homogeneous Markov model, may lead to a poor approximation in many cases. 
Recent studies~\cite{HaaslerRCK19,MOTSBP,Chen01} have shown that the inference (filtering) with population-level aggregated observations, is equivalent to an entropy regularized structured multi-marginal optimal transport (SMOT) problem. 
%CGMs can be reformulated under a graphical-structured multi-marginal optimal transport framework. 
In particular, the SMOT framework enables us to readily apply Sinkhorn algorithm to perform efficient marginal inference in collective hidden Markov models with guaranteed convergence. Following this success, \citet{Chen02} developed an approximate expectation-maximization (EM) algorithm to learn the transition parameter of a time-homogeneous hidden Markov process from the aggregated observations of population flows. Despite being simple and tractable, this new method still strictly assumes that each individual follows the time-homogeneous Markov chain, which thus may lead to a poor estimation of the true transition flows in real-world data. 

In this work, we propose to estimate the latent transition flows from aggregated data by learning the cost functions of the structured multi-marginal optimal transport framework. By doing so, our method allows the estimated transition parameters to be time-varying, and thus demonstrated improved accuracy in analyzing real-world population flows, compared with time-homogeneous Markov models. 
In particular, the main contributions of this paper are:
%(1) we propose an expectation-maximization (EM)-type algorithm  to simultaneously learn the cost functions of the formulated SMOT problem, and to estimate the transition flows using Sinkhorn belief propagation algorithm (Sec.4). The uniqueness of the recovered cost functions can be ensured under some mild conditions, as proved by the recent studies in inverse optimal transport~\cite{HaodongSun}. (2) We also investigate regularized convex optimization algorithms~\cite{SISTA} to construct cost functions as sparse linear combinations of some basis distance functions, which allow to learn more complicated cost functions than symmetric ones. (3) Experiments are conducted on both simulated and real flow data, to demonstrate the improved performance of the proposed methods in estimating latent transition flows, compared with previous related methods. 
\begin{itemize}
  \item We propose an expectation-maximization (EM)-type algorithm  to simultaneously learn the cost functions of the formulated SMOT problem, and to estimate the transition flows using Sinkhorn belief propagation algorithm (Sec.4). The uniqueness of the recovered cost functions can be ensured under some mild conditions, as proved by the recent studies in inverse optimal transport~\cite{HaodongSun}.
  \item We also investigate regularized convex optimization algorithms~\cite{SISTA} to construct cost functions as sparse linear combinations of some basis distance functions, which allow to learn more complicated cost functions than symmetric ones.
  \item Experiments are conducted on both simulated and real flow data, to demonstrate the improved performance of the proposed methods in estimating latent transition flows, compared with previous related methods.
\end{itemize}

\section{Related Work}
% For learning and estimating collective dynamics with aggregate data, prior works~\cite{} focused on the modelling of a single Markov model by maximizing aggregate posterior.
%  The recent studies in learning Markov models from aggregate data include~\cite{} with applications on tourist and migration flow analysis.
 
 Collective graphical models (CGMs) is proposed by~\cite{CGMs} as a formalism to perform inference in aggregate noisy data including ensemble flows. \citet{ApproxCGMs} studied the intractability of the exact marginal inference in CGMs, and proposed an approximate maximum a posteriori (MAP) estimation as a substitute. Following this success, \citet{MPCGMs} developed the non-linear belief propagation algorithm to perform approximate MAP inference in CGMs. Bethe-RDA is another algorithm dedicated to aggregate inference in CGMs via regularized dual averaging (RDA) with guaranteed convergence. Recently, \citet{Bernstein16} developed an approach of moments estimator to learn the parameters of the Markov model from aggregate noisy flows. 
 \citet{MOTPGM,HaaslerRCK19} recently investigated the problems of estimating ensemble flows from a graphically structured multi-marginal optimal transport perspective. In particular, \citet{MOTSBP} studied a tree-structured multi-marginal optimal transport, which allows to consider various related problems such as information fusion under a unified MOT framework. \citet{Chen01} first studied the inference (filtering) problems in CGMs based upon the tree-structured MOT framework. \citet{Chen02} derived an approximate EM algorithm to conduct learning and inference in time-homogeneous collective hidden Markov models. 
 %Nonetheless, the approximate EM scheme still admits expensive time complexity when it faces with large-scale real-world applications involving tens of thousands of states.
% To address this limitation, we are motivated to leverage the sparse network structure among the states of the Markov model, and resorts to a method of learning graph-structured transition matrix. The developed EM algorithm is appealing for large-scale settings, compared with the approximate EM algorithm~\cite{Chen02} as demonstrated in the experiments (Sec.6).

To the best of our knowledge, most of the collective graphical models assume the observed flow data are generated by time-homogeneous Markov models. In contrast, we aim to learn the time-dependent transition matrices indirectly by learning the corresponding cost functions. Our methods are based upon graphical-structured multi-marginal optimal transport (MOT) formalism. In particular, this work studies the estimation of transition flows by \emph{learning} cost functions of MOT, while the previous work~\cite{MOTPGM,MOTSBP} focus on the inference (filtering) problems with predetermined cost functions. In addition, the proposed methods are closely related to inverse optimal transport~\cite{RuilinLi,HaodongSun,SISTA}, where they aim to learn the cost functions from the observed matching, while this work focuses on estimating transition flows from marginally aggregated observations. 
Other related studies include collective flow diffusion models (CFDM)~\cite{CFDM01,CFDM02}, which can incorporate people's travel duration between locations for estimating transition flows. Neural collective graphical models (CGMs) can estimate population flows by incorporating additional spatiotemporal informtion into transition kernel parameterized by neural nets~\cite{NCGMs}. The CFDM and Neural CGMs need to explicitly model observation noise, while the proposed methods can implicitly capture noisy observations via constrained marginals.
\section{Background}
% In this section, we first introduce collective graphical models as the main language to state the problem, and then discuss the multi-marginal optimal transport.
%
\noindent\textbf{Notations.} 
By $\exp(\cdot), \ln(\cdot), \odot, /$, we denote the element-wise exponential, logarithm, multiplication, and division of vectors, matrices and tensors, respectively. The outer product is denoted by $\otimes$. 
Let $\mathbf{p}$ and $\mathbf{q}$ be two nonnegative vectors, matrices or
tensors of the same dimension. The normalized Kullback-Leibler (KL) divergence of $\mathbf{p}$
from $\mathbf{q}$ is defined as $H(\mathbf{p|q})\equiv\sum_i (p_i \ln(\frac{p_i}{q_i})-p_i + q_i)$, where $0\ln 0$ is defined to be $0$. Similarly, defined $H(\mathbf{p})\equiv H(\mathbf{p | 1}) = \sum_i (p_i \ln(p_i)-p_i+1)$, which is effectively the negative of the entropy of $\mathbf{p}$.

\subsection{Optimal transport}

Here we only consider the discrete optimal transport problems, and refer to~\citep{TOPT} for its continuous counterpart. Let  $\bm{\mu}_1\in\mathbb{R}_{\ge 0}^{d_1}$ and $\bm{\mu}_{2}\in\mathbb{R}_{\ge 0}^{d_2}$ be two distributions with equal mass. The optimal transport (OT) aims at finding a transport mapping from $\bm{\mu}_1$ to $\bm{\mu}_2$, while minimizing the total transport cost. In particular, the transport cost is defined by an underlying cost matrix
%\footnote{Note that the nonnegative constraints are imposed over the cost function for its physical meaning in the context, although the cost function is allowed to be real-valued, in \citep{TOPT}} 
$C\in \mathbb{R}^{d_1\times d_2}$, where $C_{i_1,i_2}$ measures the cost of moving an unit mass from location ${i_1}$ to ${i_2}$. Hence, the Monge-Kantorovich formulation of OT is to find a transport plan by solving the following optimization problem
\begin{align}
    \min_{M \in \Pi(\bm{\mu}_1,\bm{\mu}_2)} \langle C,M\rangle,\notag
\end{align}
where $\langle C, M\rangle = \sum_{i_1,i_2} C_{i_1,i_2}M_{i_1,i_2}$, and $\Pi(\bm{\mu}_1,\bm{\mu}_2)$ denotes the set of nonnegative matrices satisfying maringal constraints specified by $\bm{\mu}_1$ and $\bm{\mu}_2$.
Computing the exact OT problem requires solving a linear program with time complexity $\mathcal{O}(n^3\ln n)$~\citep{Pele2009FastAR}, which is too expensive for large-scale settings. To avoid excessive computational cost, \citet{EOT} introduces an entropy regularization term %$H(M) = \sum_{i_1,i_2} M_{i_1,i_2}\big(\ln M_{i_1,i_2}-1\big)$ 
$H({M}) = \sum_{i_1,i_2} (M_{i_1,i_2} \ln(M_{i_1,i_2})-M_{i_1,i_2}+1)$, and thus forms an approximate OT problem as
\begin{align}
    \min_{M \in \Pi(\bm{\mu}_1,\bm{\mu}_2)} \Big\{\langle C,M\rangle + \epsilon H(M) \Big\},\label{EOT}
\end{align}
where $\epsilon\geq0$. When $\epsilon$ approaches $0$, one recovers the canonical OT. For $\epsilon>0$, taking the dual of the approximation leads to a strictly convex optimization problem, which enables us to obtain an unique solution up to multiplication/division by a constant~\citep{franklin1989scaling}.

\subsection{Multi-marginal optimal transport}

Multi-marginal optimal transport (MOT) generalizes bi-marginal OT by considering optimal transport problems involving multiple marginal constraints. More specifically, the MOT problem is to find a transport plan between a set of marginals $\{\bm{\mu}_j\}_{j=1,2,\ldots,J}$. In this setting, the transport cost is encoded as $C = [C_{i_1,i_2,\ldots,i_J}]\in\mathbb{R}^{d_1\times\cdots\times d_J}$, 
 and the transport plan is denoted by $M = [M_{i_1,i_2,\ldots,i_J}]\in\mathbb{R}^{d_1\times\cdots\times d_J}_{\geq 0}$. For a tuple $(i_1,i_2,\ldots,i_J)$, $C_{i_1,i_2,\ldots,i_J}$ denotes the transport cost of moving an unit mass, and $M_{i_1,i_2,\ldots,i_J}$ describes the amount of mass transported for that tuple. Naturally, the Monge-Kantorovich formulation of MOT reads
 \begin{align}
    \min_{M} \quad &\langle C,M\rangle\label{MOT}\\%_{M \in \Pi(\mu_1,\ldots,\mu_J)}
    \text{subject to}\quad&
    P_j(M) = \bm{\mu}_j,\quad \text{for}\ j \in \Gamma,\nonumber
\end{align}
where $\Gamma \subset \{1,2,\ldots,J\}$ denotes an index set specifying which marginal constraints are given. The projection of the tensor $M$ on its $j$-th marginal is given by
 \begin{align}%
    P_j(M) = \sum_{i_1, \ldots,i_{j-1},i_{j+1},\ldots, i_J} M_{i_1,\ldots,i_{j-1},i_j,i_{j+1},\ldots,i_J}.\label{Proj}
\end{align}

Note that the original multi-marginal optimal transport formulation~\citep{Pass11,Pass12} specifies all the marginal distributions as its constraints. Here we consider the case where only a subset of marginals are explicitly given, i.e., $\Gamma \subset \{1,2,\ldots,J\}$. 
This arises in many cases of interests including dynamic network flows~\citep{haasler2021scalable} and Barycenter problems~\citep{IBProj}. 
% We also note that bi-marginal OT is a special case of MOT by letting $J=2$ and $\Gamma = \{1,2\}$. 
% The MOT problem can be solved by a linear program with a supercubic time complexity. As we did for bi-marginal OT, this computational burden can be partly alleviated by introducing entropy regularization to form an approximate MOT as

The entropy regularized MOT reads
 \begin{align*}
    \min_{M} \quad &\Big\{\langle C,M\rangle + \epsilon H(M)\Big\}\\%_{M \in \Pi(\mu_1,\ldots,\mu_J)}
    \text{subject to}\quad&
    P_j(M) = \bm{\mu}_j,\quad \text{for}\ j \in \Gamma.
    % \text{s.t.}\quad&
    % \sum_{i_1, \ldots,i_{j-1},i_{j+1}, i_J} \pi(i_1, \ldots, i_J) = \mu_j,\quad \text{for}\ j = 1,\ldots,J,
\end{align*}

Using the Lagrangian duality theory, it is not hard to see the optimal solution of the entropy regularized MOT is of the form
 %\begin{align*}
 $M = K\odot B$
%\end{align*}
where $K = \exp(-C/\epsilon)$, and\\ $B = \mathbf{b}_1 \otimes \cdots \otimes \mathbf{b}_J$ with
 \begin{align*}
\mathbf{b}_j = \begin{cases} \exp(\bm{\alpha}_j/\epsilon), & \text{if}\ j\in\Gamma  \notag\\ \mathbf{1}, & \text{otherwise}\end{cases} 
\end{align*}
where $\bm{\alpha}_j\in\mathbb{R}^n$ denotes the dual variable corresponding to the constraint $P_j(M) = \bm{\mu}_j$, for $j\in\Gamma$. The generalized Sinkhorn algorithm solves entropy regularized MOT problems by iteratively updating the vectors $\mathbf{b}_j$, for $j\in\Gamma$, as
 \begin{align*}
\mathbf{b}_j \leftarrow \mathbf{b}_j\odot\bm{\mu}_j/P_j(K\odot B).
\end{align*}
% The Sinkhorn algorithm can be derived as Bregman iterations~\citep{IBProj}, or dual block coordinate ascent~\citep{GSinkhorn}, and thus admits a global convergence guarantee with linear rate~\citep{ZQL}. 

Note that the computational complexity of Sinkhorn algorithm still scales exponentially with $J$ because the number of elements in $M$ is $d_1\times\cdots\times d_J$.

 Fortunately, the tree-structured cost tensors in many cases of interests, allow us to make the computation of the marginal projections feasible~\citep{MOTPGM}. More specifically, let $\mathcal{T} = (\mathcal{V},\mathcal{E})$ be a tree with $\mathcal{V}$ denoting the nodes, and $\mathcal{E}$ the edges. Assume that the cost tensor $C$ can be decomposed according to a tree structure $\mathcal{T}=(\mathcal{V},\mathcal{E})$ with $J$ nodes as
  \begin{align*}
C_{i_1,\ldots,i_J} = \sum_{(v,u)\in \mathcal{E}} C^{(v,u)}_{i_{v},i_{u}},
\end{align*}
where $C^{(v,u)}$ denotes the cost matrix between marginals $\bm{\mu}_v$ and $\bm{\mu}_u$, for $(v,u)\in \mathcal{E}$.  
By letting $K^{(v,u)}\equiv\exp(-C^{(v,u)}/\epsilon)$, 
% the transport plan is thus of the form
%   \begin{align*}
%  M = K\odot B = \prod_{(v,u)\in E}K^{(v,u)} \prod_{j\in V}\mathbf{b}_j.
% \end{align*}
% Analogously, 
the projection of $M$ on the $j$-th marginal is specified as
  \begin{align}
 &P_j(M)_{i_j}\\%\label{BPProj}\\
%  &= \sum_{i_1, \ldots,i_{j-1},i_{j+1}, i_J}\prod_{(v,u)\in E}K^{(v,u)}_{i_v,i_u} \prod_{v\in V}\mathbf({b}_v)_{i_v}\\
%  & 
 &= (\mathbf{b}_j)_{i_j}\sum_{i_1, \ldots,i_{j-1},i_{j+1},\ldots, i_J}\prod_{(v,u)\in \mathcal{E}}K^{(v,u)}_{i_v,i_u} \prod_{v\in \mathcal{V}\backslash j}(\mathbf{b}_v)_{i_v}.\notag
\end{align}
This sum only involves matrix-vector multiplications, and hence substantially reduces the computational complexity compared with the brute force summation in Eq.~\ref{Proj}. 
% We remark that the entropy regularized MOT with a tree-structured cost, is equivalent to the constrained marginal inference in PGMs.  For the justifications about this connection, we refer the readers to ~\citep{MOTPGM}. % with additional marginal constraints over a subset of variable nodes
% Consequently, sum product belief propagation algorithms, which were originally designed for marginal inference in PGMs, can be readily combined with Sinkhorn algorithm, to perform the marginal projections in Eq.~(\ref{BPProj}). 
% Hence, t
The full algorithm is introduced as Sinkhorn belief propagation algorithm  in~\citep{MOTPGM}, for graphically structured MOT problems. 
\begin{figure}[t]
  \centering
      \includegraphics[width=12.5cm,height=9cm, keepaspectratio]{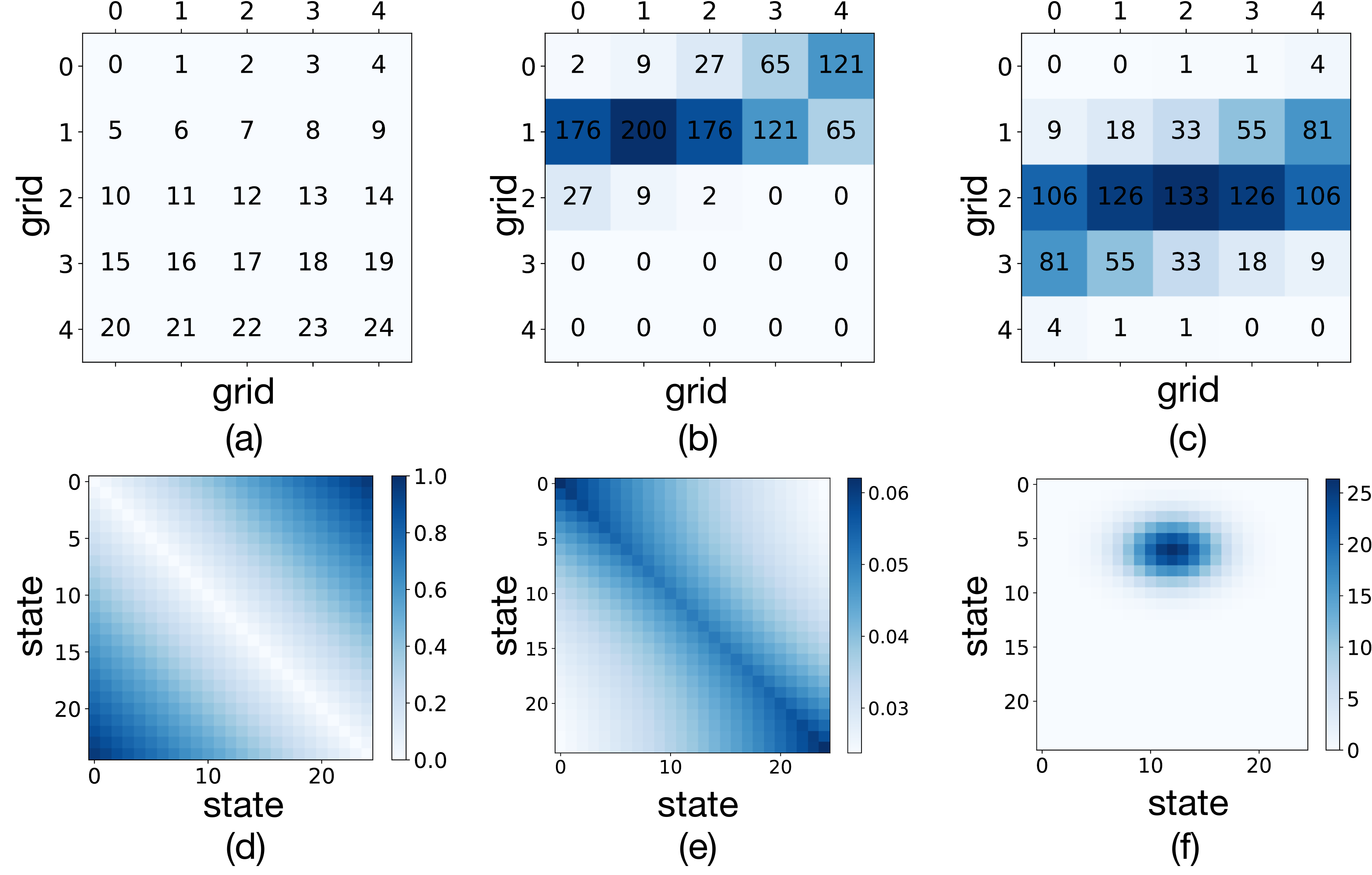}
      %\vspace{-0.51em}
\caption{An illustration of the studied problem. The $5\times5$ grid cells form 25 states (a). There are $1,000$ individuals moving among these states. The aggregated observations (the number in each cell, indicates the observed count of individuals in that state) are measured at two consecutive time steps, as shown in (b) and (c), respectively. This work aims to estimate the latent transition flow (f) (the value of $(i,j)$-th entry denotes the number of individuals moving from state $i$ to state $j$ at the target time step)  by learning the cost matrices (d). The transition matrix is displayed in (e).}
\label{peopleflow}
\end{figure}

% %\input{section_inference}
% \linespread{0.86}
%\input{section_problem_formulation}
\section{Problem Formulation}
Consider a population of $N$ individuals (e.g., pedestrians, bikes, cars), each of which independently behaves according to a Markov chain. Let the states of the Markov chain be $X = \{X_1,\ldots,X_S\}$ with $S$ being the number of states. In particular, the transition parameters of the Markov chain allows to be time-varying, and specified by $\mathbf{A}^{t}$, where\\ ${A}_{ij}^{t} = p(s_{t+1}=x_j\mid s_t = x_i)$ denotes the transition probability from state $x_i$ at time $t$, to state $x_j$ at time $t+1$. Let $(\bm{\mu}_t)_i$ denote the number of individuals appearing in state $x_i$ at time $t$, and $\mathbf{M}^t=[M_{ij}^t]$, where $M^t_{ij}$ denote the number of individuals moving from state $x_i$ at time $t$, to state $x_j$ at time $t+1$. Fig.~\ref{peopleflow} illustrates an example of the studied problem.
 The probability of the transition flow observed during the time interval $[t,t+1]$ is given by
\begin{align}
    p(\mathbf{M}^t) = \prod_{i=1}^S\frac{(\bm{\mu}_t)_i}{\prod_{j=1}^S M^t_{ij}}\prod_{j=1}^S(A_{ij}^t)^{M_{ij}^t}.\nonumber
\end{align}

Interestingly, a large deviation interpretation~\cite{MOTSBP} has shown that as the number of individuals $N$ tends to infinity, if $\frac{1}{N}\bm{\mu}_{t}\rightarrow\bar{\bm\mu}_{t}$, and $\frac{1}{N}\mathbf{M}^{t}\rightarrow\bar{\mathbf{M}}^{t}$, the log-likelihood of the transition flow $\mathbf{M}^{t}$ can be well approximated as
\begin{align}
    \frac{1}{N}\log p({\mathbf{M}}^t) \rightarrow -H(\bar{\mathbf{M}}^{t}\mid \text{diag}(\bar{\bm\mu}_{t})\mathbf{A}^{t}).\nonumber
\end{align}
%where $H(p|q)\equiv \sum_i p_i\log(p_i/q_i)$ denotes the Kullback-Leibler (KL) divergence of $p$ from $q$.

%%%%%%%%%%%%%%%%%%%%%%%%%%%%%%%%%
% Given the aggregated observations $\{\bm{\mu}_j\}_{j\in\Gamma}$\footnote{Hereafter, we use $\bm{\mu}_t$, ${\mathbf{M}}^{t}$ to denote the normalized observations $\bar{\bm\mu}_t,\bar{\mathbf{M}}^{t}$, respectively, for ease of notation. 
% % The estimated flow can be recovered from the estimation by $\hat{\mathbf{M}}^{t} = N\bar{\mathbf{M}}^{t}$.
% }, the problem of estimating transition flows $\{\mathbf{M}^{t}\}_{t=1}^{T-1}$ can be naturally reformulated as a convex optimization problem given by

% \begin{align}
% \min_{\left\{{\mathbf{M}}^{[1:(T-1)]}\right\}} \quad& \sum_{t=1}^{T-1}H({\mathbf{M}}^{t}\mid \text{diag}(\bm{\mu}_{t})\mathbf{A}^{t}) %F_{\mathrm{B}}(\mathbf{n},\mathbf{y};\bm{\theta}) 
% \label{BFE}\\
% \text{subject to}\ 
%     & {\mathbf{M}}^{t}\mathbf{1} = \bm{\mu}_{t},\nonumber\\
%  & ({\mathbf{M}}^{t})^{\mathrm T}\mathbf{1} = \bm{\mu}_{t+1},\quad \text{for}\ \ t =1,\ldots,T-1.\nonumber
% \end{align}
%%%%%%%%%%%%%%%%%%%%%%%%%%%
\begin{figure}%[t]
  \centering
      \includegraphics[width=6cm,height=10cm, keepaspectratio]{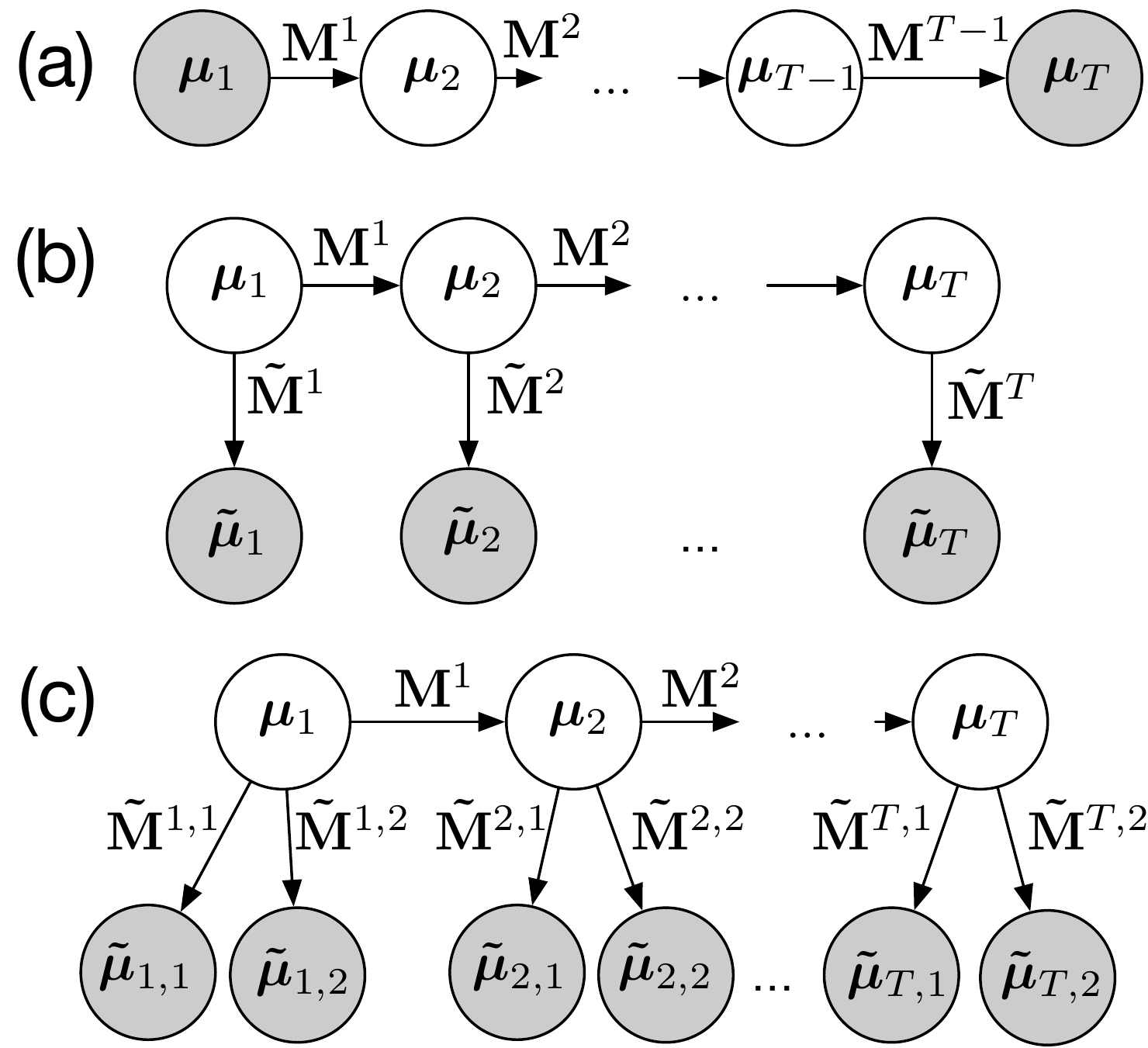}
       \caption{An illustration of the original ensemble flow estimation problem in~\cite{MOTSBP,MOTPGM}, where the transition parameter $\mathbf{A}$ and two marginals $\bm{\mu}_1$ and $\bm{\mu}_T$ are known, and the goal is to estimate transition flows and intermediate marginals (a). Latent flow estimation problems (b) and (c), provide multiple noisy marginals, for which we aim to estimate the underlying transition flows, marginals, and to learn transition parameters. In (c), $\bm{\tilde \mu}_{t,s}$ denotes the $s$-th noisy measurements at time $t$, and $\mathbf{\tilde M}^{t,s}$ denotes the transition flows between $\bm{\tilde \mu}_{t,s}$ and $\bm{\mu}_{t}$.}
\label{TreeGraphs}
\end{figure}

As shown in Fig.~\ref{TreeGraphs}(b), given the noisy aggregated observations $\{\bm{\tilde \mu}_t\}_{t = 1}^T$
\footnote{Hereafter, we use $\bm{\mu}_t$, ${\mathbf{M}}^{t}$ to denote the normalized observations $\bar{\bm\mu}_t,\bar{\mathbf{M}}^{t}$, respectively, for ease of notation.}, the problem of estimating latent ensemble flows can be naturally reformulated as a convex optimization problem given by

\begin{align}
\mathop{\rm{min}}_{{\mathbf{M}}^{[1:(T-1)]},\atop {\mathbf{\tilde M}}^{[1:(T-1)]}, \bm{\mu}_{[1:T]}} \quad& \left\{\sum_{t=1}^{T-1}H({\mathbf{M}}^{t}\mid \text{diag}(\bm{\mu}_{t})\mathbf{A}^{t})\right.  \left.+ \sum_{t=1}^{T-1}H({\mathbf{\tilde M}}^{t}\mid \text{diag}(\bm{\tilde \mu}_{t})\mathbf{\tilde A}) \right\}
%F_{\mathrm{B}}(\mathbf{n},\mathbf{y};\bm{\theta}) 
\label{BFE}\\
\text{subject to}\
& {\mathbf{\tilde M}}^{t}\mathbf{1} = \bm{\mu}_{t},\quad ({\mathbf{\tilde M}}^{t})^{\mathrm T}\mathbf{1} = \bm{\tilde \mu}_{t},\nonumber\\&{\mathbf{M}}^{t}\mathbf{1} = \bm{\mu}_{t},\quad ({\mathbf{M}}^{t})^{\mathrm T}\mathbf{1} = \bm{\mu}_{t+1},\nonumber\\& \text{for}\ \ t =1,\ldots,T-1,\nonumber
\end{align}
where $\mathbf{\tilde A}$ denotes the emission parameter that determines the conditional distribution of the noisy observation $\bm{\tilde \mu}_{t}$ given the true marginal $\bm{\mu}_{t}$, and $\mathbf{\tilde M}^{t}$ refers to the transition flow between the true marginal $\bm{\mu}_{t}$ and the noisy observation $\bm{\tilde \mu}_{t}$.

%%%%%%%%%%%%%%%%%%%%%%%%%%%
\noindent\textbf{Remark 1.} 
% The aforementioned problem naturally induces a tree-structure with each edge associated with a time-dependent transition probability matrix $\mathbf{A}^{t}$. 
If we define the cost matrix $\mathbf{C}^t=-\epsilon\log(\mathbf{A}^{t})$ and $\mathbf{\tilde C}=-\epsilon\log(\mathbf{\tilde A})$, the convex optimization problem in Eq.~{\ref{BFE}} is equivalent to a multi-marginal optimal transport problem specified by

% \begin{align}
% \min_{\{\mathbf{A}^{t},\bar{\mathbf{M}}_{t}\}} \quad& \sum_{t=1}^{T-1}H(\bar{\mathbf{M}}_{t}\mid \text{diag}(\bar{\mu}_{t})\mathbf{A}^{t}) %F_{\mathrm{B}}(\mathbf{n},\mathbf{y};\bm{\theta}) 
% \label{BFE}\\
% \text{subject to}\ 
%     & \bar{\mathbf{M}}_{t}\mathbf{1} = \bar{\mu}_{t},\nonumber\\
%  & (\bar{\mathbf{M}}_{t})^{\mathrm T}\mathbf{1} = \bar{\mu}_{t+1},\quad \text{for}\ \ t =1,\ldots,T-1.\nonumber
% \end{align}
 \begin{align}
    \min_{\mathbf{M}} \quad &\Big\{\langle \mathbf{C},\mathbf{M}\rangle + \epsilon H(\mathbf{M}\mid \mathbf{1}_{S\times\cdots\times S})\Big\}\label{EqvMOT}\\%_{M \in \Pi(\mu_1,\ldots,\mu_J)}
    \text{subject to}\quad&
    P_j(\mathbf{M}) = \bm{\mu}_j,\quad \text{for}\ j \in \Gamma,\nonumber
    % \text{s.t.}\quad&
    % \sum_{i_1, \ldots,i_{j-1},i_{j+1}, i_J} \pi(i_1, \ldots, i_J) = \mu_j,\quad \text{for}\ j = 1,\ldots,J,
\end{align}
where the cost tensor $\mathbf{C}\in \mathrm{R}^{S\times\cdots\times S}$ decomposes as $\mathbf{C}_{i_1,\ldots,i_J} = \sum_{(u,v)\in E}\mathbf{C}^{(u,v)}_{i_u,i_v}$, according to the tree structure in Fig.~\ref{TreeGraphs}(b). The marginal observation $\bm\mu_t$ equals to the projection of the tensor-valued transport plan $\mathbf{M}\in \mathrm{R}^{S\times\cdots\times S}$ on its $t$-th mode. Similarly, the transition flow $\mathbf{M}^t$ can be obtained by the projection of the tensor-valued transport plan $\mathbf{M}$ on its $(t,t+1)$-th modes, i.e., $\mathbf{M}^t=P_{(t,t+1)}(\mathbf{M})$. We refer to Sec.4.2 in~\cite{MOTSBP} for the detailed proof of the equivalence between Eq.~\ref{BFE} and~\ref{EqvMOT}.

As illustrated in Fig.~\ref{TreeGraphs}(a), the multi-marginal optimal transport based methods~\cite{MOTSBP,MOTPGM}, aim to estimate the transition flow matrices $\{\mathbf{M}^t\}_{t=1}^{T-1}$ given the two marginal observations $\bm{\mu}_1$ and $\bm{\mu}_T$, using the predetermined transition matrix $\mathbf{A}$, for time-homogeneous Markov models. Fig.~\ref{TreeGraphs}(b) illustrates a scenario in which the multiple noisy marginals are accessible. In particular, our goal is to simultaneously recover the transition flows $\{\mathbf{M}^t\}_{t=1}^{T-1}$ and to learn the unknown cost matrices $\{\mathbf{C}^t\}_{t=1}^{T-1}$ based upon the multiple marginal observations $\bm{\tilde \mu}_{t}, \text{for}\ \ t =1,\ldots,T$.  
To this end, an EM-type algorithm is developed to solve the problem of estimating transition flows. More specifically,  in the M-step, we consider learning the cost matrices $\mathbf{C}^{t(\ell)}$ of $\ell$-th iteration given the two marginal observations $\bm{\mu}_t$ and $\bm{\mu}_{t+1}$, and the estimated transition flow $\mathbf{M}^{t(\ell)}$ of the  previous iteration. In the E-step, the expectation of transition flow $\mathbf{M}^{t (\ell+1)}$ is updated based upon $\mathbf{C}^{t (\ell)}$. Moreover, via the tree-structured MOT framework, the proposed method can be well extended to more complicated scenarios where multiple noisy aggregated observations are available for each marginal observation (Fig.~\ref{TreeGraphs}(c)).

\noindent\textbf{E-step.}
With the cost matrices $\{\mathbf{C}^{t}\}$ updated in the M-step, the optimization problem in Eq.{\ref{BFE}} can be equivalently solved via an entropy regularized multi-marginal optimal transport formulation in Eq.~\ref{EqvMOT}. 
%  \begin{align*}
%     \min_{\mathbf{M}} \quad &\Big\{\langle \mathbf{C},\mathbf{M}\rangle - \epsilon H(\mathbf{M}\mid \mathbf{1}_{S\times,\ldots,S}) \Big\}\\%_{M \in \Pi(\mu_1,\ldots,\mu_J)}
%     \text{subject to}\quad&
%     P_j(\mathbf{M}) = \bm{\mu}_j,\quad \text{for}\ j \in \Gamma.
%     % \text{s.t.}\quad&
%     % \sum_{i_1, \ldots,i_{j-1},i_{j+1}, i_J} \pi(i_1, \ldots, i_J) = \mu_j,\quad \text{for}\ j = 1,\ldots,J.
% \end{align*}
Hence, the tree-structure induced by the latent flow estimation, enables us to readily utilize Sinkhorn belief propagation (SBP) algorithm to recover the latent transition flows $\{\mathbf{M}^{t}\}_{t=1}^{T-1}$. The SBP algorithm for the E-step is detailed in Algorithm~\ref{alg_SBP}.

\begin{algorithm}[t]%\SetAlgoNoLine
\caption{Sinkhorn Belief Propagation Algorithm}\label{alg_SBP}
\begin{algorithmic}[1]
\REQUIRE Tree-structured graph $\mathcal{T}=(\mathcal{V},\mathcal{E})$ with $\mathcal{V}$ the node set, $\mathcal{E}$ the edge set, and the indices of the constrained marginals $\Gamma$, marginal observations $\bm{\tilde \mu}_{t}$ and edge potentials $\bm{\Phi}^{t} = \exp(-\mathbf{C}^t/\epsilon)\ \text{for}\ \ t =1,\ldots,T$%events data $\mathcal{D}=\{(t_i,s_i,d_i)\}_{i=1}^N$, $\{\Phi$, $\Omega\}$ inferred by the HGaP-EPM, time scale $\delta$%, maximum iterations $\mathcal{J}$ 
\ENSURE 
the transition flows \\
$\mathbf{M}^t(x_t,x_v)$\\
$\propto\bm{\phi}^t(x_t,x_v)\prod\limits_{\substack{k\in N(t)}}{\mathbf{\breve{M}}_{k\rightarrow t}(x_t)}\prod\limits_{\substack{k\in N(v)}}{\mathbf{\breve{M}}_{k\rightarrow v}(x_v)}
$
% the node contingency tables \\$\mathbf{n}_v(x_v)\propto \prod\limits_{\substack{k\in N(v)}}{\mathbf{{m}}_{k\rightarrow v}(x_v)},\quad\forall v\notin\Gamma$%$\{\mu_{u,k,k',v}\}$, $\{\alpha_{kk'}\}$%, $\{(z_i^s, z_i^d)\}$
\STATE Initialize the messages $\mathbf{\breve{M}}_{v\rightarrow u}(x_u),\ \forall (v,u)\in \mathcal{E}$\\
\REPEAT %FOR{$l$ = 1:$\mathcal{J}$}
	\FOR{$v\in{\Gamma}$}
	\STATE Update $\mathbf{\breve{M}}_{v\rightarrow u}(x_u)\propto\sum\limits_{\substack{x_v}}\phi^{v}(x_v,x_u)\frac{\bm{\mu}_v(x_v)}{\mathbf{\breve{M}}_{u\rightarrow v}(x_v)}$,\\$\forall u \in N(v)$%(Eq.~\ref{em_p})$
	\STATE Update all the messages on the path from $v$ to $v_{\mathrm{next}}$\\ $\mathbf{\breve{M}}_{v\rightarrow u}(x_u)\propto\sum\limits_{\substack{x_v}}\phi^{v}(x_v,x_u)\prod\limits_{\substack{k\in N(v)\backslash u}}{\mathbf{\breve{M}}_{k\rightarrow v}(x_v)}$
% 	$\sum_{\substack{i=0 \\ i\neq 4}}^n i$, or
% $\sum\limits_{\substack{i=0 \\ i\neq 4}}^n i$
% 	\STATE Update the intensity function ${\lambda_{s_i,d_i}(t_i)}$ (Eq.~\ref{intensity})%(Eqs.~\ref{eq_z};~\ref{eq:lcount})\label{intensity}
	\ENDFOR
% 	\STATE  Update $\widehat{m}_{u,k,k',v}$ and $\widecheck{m}_{u,k,k',v}$ (Eq.~\ref{em_m})
% 	\STATE  Update the base intensities $\{\mu_{u,k,k',v}\}$ (Eq.~\ref{em_mu})
% 	\STATE  Update the parameters $\{\bm{\beta}_{kk'}\}$, $\{\omega_{u,k,k',v}\}$, $\{ \psi_{k,k'}\}$ (Eqs.~\ref{em_beta};~\ref{em_omega};~\ref{em_psi})
% 	\STATE  Update the kernel parameters $\{\alpha_{k,k'}\}$  (Eq.~\ref{em_alpha})
\UNTIL {convergence}
\end{algorithmic}
\end{algorithm}

\noindent\textbf{M-step.}
Given the marginal observations $\{\bm{\mu}_t\}_{t=1}^T$,
 and the estimated transition flows $\{\mathbf{M}_{t}\}_{t=1}^{T-1}$, the parameter learning of the collective graphical models in Eq.~\ref{BFE}, becomes an inverse multi-marginal optimal transport problem given by
%  \begin{align*}
%     \min_{\mathbf{C}} \quad & \epsilon H(\mathbf{M}^{(\ell)}\mid \mathbf{M}^c) + R(\mathbf{C})\\%_{M \in \Pi(\mu_1,\ldots,\mu_J)}
%     \text{subject to}\quad&
%     \mathbf{M}^c\equiv \arg\min_{\mathbf{M}\in\Pi(\mu_1,\dots,\mu_T)} \langle \mathbf{C},\mathbf{M}\rangle - \epsilon H(\mathbf{M}\mid \mathbf{1}_{K\times,\ldots,K}).
%     % P_j(\mathbf{M}) = \bm{\mu}_j,\quad \text{for}\ j \in \Gamma,
% \end{align*}

\begin{equation}
    \min_{\mathbf{C},\bm{\alpha}}\ \Big\{F(\bm{\alpha},\mathbf{C}) + R(\mathbf{C})\Big\},\label{IMOT0}
\end{equation}
where %$F(\bm{\alpha},\mathbf{C})\equiv\langle \mathbf{M}^{(\ell)}, \mathbf{C} \rangle - \sum_{t} \langle \bm{\alpha_t}, \bm{\mu}_t\rangle + \epsilon \langle e^{(\sum_t\bm{\alpha}_t - \mathbf{C})/\epsilon}, \mathbf{1}\rangle$ 
$F(\bm{\alpha},\mathbf{C})\equiv\langle \mathbf{M}^{(\ell)}, \mathbf{C} \rangle - \sum_{t} \langle \bm{\alpha_t}, \bm{\mu}_t\rangle + \epsilon \langle \mathbf{K}, \mathbf{B}\rangle$ is a convex function,  $\bm{\alpha}\equiv[\bm{\alpha}_1,\ldots,\bm{\alpha}_T]$ denote the dual variables corresponding to the marginal constraints $\{P_t(\mathbf{M}) = \bm{\mu}_t\}_{t=1}^T$, $\mathbf{K}\equiv\exp(-\frac{\mathbf{C}}{\epsilon})$, $\mathbf{B}\equiv\mathbf{b}_1\otimes\cdots\otimes\mathbf{b}_T$ where $\mathbf{b}_t\equiv\exp(\frac{\bm{\alpha}_t}{\epsilon})$, and $R(\mathbf{C})$ is the regularization imposed on the cost tensor $\mathbf{C}$. The derivation of Eq.~\ref{IMOT0} is detailed in the appendix. Note that the optimization problem in Eq.{\ref{IMOT0}} admits infinitely many solutions without additional regularization imposing on cost tensor $\mathbf{C}$. 

Some recent advancements~\cite{RuilinLi,HaodongSun} in solving the problem of inverse optimal transport(IOT), has proved that the IOT problem admits an unique solution if %$\mathcal{C}$ is the 
the cost function is restricted to belong to a set of symmetric matrices with zero diagonal elements, and thus 
the proximal operator can be specified by
\begin{equation*}
\mathbf{C}^t = \mathrm{prox}_{\gamma R}(\hat{\mathbf{C}^t}) = (\hat{\mathbf{C}^t}+({\hat{\mathbf{C}}^t})^{\mathrm T})/{2},
\end{equation*}
and followed by enforcing the diagonal entries of $\mathbf{C}^t$ to be $0$. In our case, the cost tensor of the inverse multi-marginal optimal transport problem,  naturally decouples, according to the tree structure as $\mathbf{C}_{i_1,\ldots,i_J} = \sum_{(u,v)\in \mathcal{E}}\mathbf{C}^{(u,v)}_{i_u,i_v}$. Thus, we impose symmetric and zero-diagonal constraints straightforwardly on each of the cost matrices $\mathbf{C}^t$, instead of restricting a symmetric cost tensor. One instance of symmetric cost matrices is $\mathbf{C}^t=[{C}^t_{ij}]$ with ${C}^t_{ij} = |x_i-x_j|^2$ where $x_i$ and $x_j$ denote $i$-th and $j$-th locations, respectively. 
%The optimization problem in Eq.{\ref{IMOT}} reduces to the MOT problem if the cost function $\mathbf{C}$ is fixed. Hence, 
In particular, $\bm{\alpha}$ can be updated using Sinkhorn belief propagation algorithm for entropy regularized MOT. More specifically, to solve the convex optimization problem in Eq.\ref{IMOT0}, a block coordinate descent scheme detailed in Algorithm~\ref{alg_IOT} can be considered to alternatively update $\mathbf{C}$ and $\bm{\alpha}$.

\begin{algorithm}[t]%\SetAlgoNoLine
\caption{Iterative Scaling Algorithm for Learning Cost Functions}\label{alg_IOT}
\begin{algorithmic}[1]
\REQUIRE The expected transition flows $\mathbf{M}^{t (\ell)}$, and the marginal observations $\bm{\mu}_t$ and $\bm{\mu}_{t+1}$
\ENSURE the cost matrix $\mathbf{C}^t$%, and $\alpha_t=\epsilon\log({u}_t)$
\STATE Initialize $\epsilon, \mathbf{C}^t$, $\alpha^v$, and set $u^v=\exp(\alpha^v/\epsilon)$ for $v=t,t+1$
\REPEAT %FOR{$l$ = 1:$\mathcal{J}$}
\STATE ${\Sigma}\leftarrow \exp\Big(-\frac{\mathbf{C}^t}{\epsilon}\Big)$\\
\STATE ${u_t}\leftarrow \bm{\mu}_t/(\Sigma u_{t+1})$\\
\STATE ${u_{t+1}}\leftarrow \bm{\mu}_{t+1}/(\Sigma^{\mathrm T} u_{t})$\\
\STATE ${\Sigma}\leftarrow \mathbf{M}^{t (\ell)}/(u_{t}u_{t+1}^{\mathrm T})$ \\
\STATE $\mathbf{C}^t = \mathrm{prox}_{\gamma R}(-\epsilon\log(\Sigma))$\\
\UNTIL {convergence}
\end{algorithmic}
\end{algorithm}
\begin{algorithm}[t]%\SetAlgoNoLine
\caption{ISTA Algorithm for Learning Cost Functions}\label{alg_SISTA}
\begin{algorithmic}[1]
\REQUIRE The expected transition flows $\mathbf{M}^{t (\ell)}$, marginal observations $\bm{\mu}_t$, $\bm{\mu}_{t+1}$, and basis distance matrices $\{\mathbf{D}^q\}_{q=1}^Q$
\ENSURE the cost matrix $\mathbf{C}^t$%, and $\alpha_t=\epsilon\log({u}_t)$
\STATE Initialize $\epsilon, \bm{\beta}^t, \alpha^v$, and set $\mathbf{C}^t = \sum_{q=1}^{Q}\beta^t_q \mathbf{D}^q$, and $u^v=\exp(\alpha^v/\epsilon)$ for $v=t,t+1$
\REPEAT %FOR{$l$ = 1:$\mathcal{J}$}
\STATE Set $\mathbf{C}^t = \sum_{q=1}^{Q}\beta^t_q \mathbf{D}^q$\\
\STATE ${\Sigma}\leftarrow \exp\Big(-\frac{\mathbf{C}^t}{\epsilon}\Big)$\\
\STATE ${u_t}\leftarrow \bm{\mu}_t/(\Sigma u_{t+1})$\\
\STATE ${u_{t+1}}\leftarrow \bm{\mu}_{t+1}/(\Sigma^{\mathrm T} u_{t})$\\
\STATE $\mathbf{M}^{t(\beta)}\leftarrow u^t\odot \Sigma\odot u^{t+1}$\\
\STATE $\beta_k^{t (\ell+1)} = \mathrm{prox}_{\rho\gamma|\cdot|}\Big(\beta_k^{t (\ell)}-\rho\sum_{i,j}(\mathbf{M}_{ij}^{t(\ell)}-\mathbf{M}_{ij}^{t (\beta)})\Big)$\\
% \STATE ${\Sigma}\leftarrow \mathbf{M}^{t (\ell)}/(u_{t}u_{t+1}^{\mathrm T})$ \\
% \STATE $\mathbf{C}^t = \mathrm{prox}_{\gamma R}(-\epsilon\log(\Sigma))$\\
\UNTIL {convergence}
\end{algorithmic}
\end{algorithm}

Although the symmetric and zero-diagonal constraints ensure the unique solution, the cost matrices between the states might be more complex. For instance, in many urban population data~\cite{Chen02}, most individuals are transitioning from suburb towards downtown areas in the early morning, while they are moving back in the opposite direction, in the late evening. Inspired by recent advances in the optimal matching studies~\cite{SISTA}, we consider constructing the time-dependent cost matrices as a sparse, linear combination of basis distance matrices. More specifically, $\mathbf{C}^t = \sum_{q=1}^{Q}\beta^t_q \mathbf{D}^q$ where $\mathbf{D}_{ij}^q = |x_i-x_j|^q$ denotes the $(i,j)$-th element of the $q$-th basis distance matrix, $x_i$ is the $i$-th location, $\bm{\beta}^t=[\beta^t_1,\ldots,\beta^t_Q]$ is a sparse coefficient vector with $q$-th element determining the usage of $\mathbf{D}^q$ in the construction of $\mathbf{C}^t$. Thus, the learning problem of the cost matrices $\mathbf{C}^t$ reduces to an optimization problem with respect to $\bm{\beta}^t$ and $\bm{\alpha}$ as
\begin{equation*}\label{IMOT}
    \min_{\bm{\beta},\bm{\alpha}^t, \bm{\alpha}^{t+1}}\ \Big\{F(\bm{\alpha}^t, \bm{\alpha}^{t+1},\bm{\beta}) + \gamma|\bm{\beta}|_{1}\Big\},
\end{equation*}
where $F(\bm{\alpha}^t, \bm{\alpha}^{t+1},\bm{\beta})\equiv\sum_{i_t,i_{t+1}} e^{[(\bm{\alpha}^t)_{i_t}+ (\bm{\alpha}^{t+1})_{i_{t+1}} -\mathbf{C}^{t}_{i_t,i_{t+1}}]} + \sum_{i_t,i_{t+1}}[\mathbf{C}^{t}_{i_t,i_{t+1}}-(\bm{\alpha}^t)_{i_t}-(\bm{\alpha}^{t+1})_{i_{t+1}}]$, and $\bm{\alpha}^t, \bm{\alpha}^{t+1}$ denote the dual variables corresponding to $\bm{\mu}^t, \bm{\mu}^{t+1}$, respectively, and the $\ell_1$ penalty term is to enforce a sparse coefficient vector $\bm{\beta}^t$. As we did in Algorithm~\ref{alg_IOT}, $\bm\alpha$ can be updated using Sinkhorn algorithm, and $\bm\beta$ is updated using an iterative shrinkage-thresholding algorithm (ISTA)~\cite{SISTA}, which leads to the second block coordinate descent scheme as detailed in Algorithm~\ref{alg_SISTA}, for learning cost matrices. The proximal operator is given by the soft-thresholding operator specified by
\begin{align}
% \mathrm{prox}_{\rho\gamma|\cdot|}(x) = \begin{cases} x-\rho\gamma & \text{if}\ x>\rho\gamma\\
% 0 & \text{if}\ |x|\leq\rho\gamma\\
% x+\rho\gamma &\text{if}\ x<-\rho\gamma
% \end{cases}.\nonumber\\
\mathrm{prox}_{\rho\gamma|\cdot|}(x) = \mathrm{sign}(x) \mathrm{max}\{|x|-\rho\gamma, 0\}.\nonumber
\end{align}

The proposed EM-type algorithm is to estimate latent transition flows by iteratively implementing the Sinkhorn belief propagation in the E-step to estimate the expected transition flows, and to learn the cost functions using Algorithm~\ref{alg_IOT} or Algorithm~\ref{alg_SISTA}. Hereafter, we denote the two developed EM-type algorithms as Sinkhorn belief propagation inverse symmetric transport cost (SBP-ISTC), and Sinkhorn belief propagation iterative shrinkage-thresholding (SBP-ISTA) algorithms.

\noindent\textbf{Computational Cost.}
For Sinkhorn belief propagation algorithm implemented in the E-step, computing the transition flow matrices $\{\mathbf{M}^t\}_{t=1}^{T-1}$ takes $\mathcal{O}(TS^2)$ time, where $S$ is the number of states, and $T$ is the number of vertices, i.e., $T=|\mathcal{E}|$. To update the cost function in the M-step, the iterative scaling algorithms enjoy the quadratic computational complexity $\mathcal{O}(S^2)$. For Algorithm~\ref{alg_SISTA}, the computation cost of ISTA algorithm scales with $\mathcal{O}(Q\mathcal{L})$, where $Q$ denotes the number of basis distance matrices, and $\mathcal{L}$ is the number of inner iterations for the convergence of ISTA algorithm.

% 
% %\clearpage

%\input{section_model_v3}

% \input{section_relatedworks}

%\input{section_experiments}

%\begin{table*}[htbp]
% \centering
%\caption{Test table.}
%\label{tbl:test}
%%\caption{Normalized mean absolute error (NAE) for the estimation of transition flows in the real-world datasets}
%%\label{realdata}
% \begin{tabular}{ c|c|c|c|c } \hline
%%  & t1 & t2 & t3 & t4 \\ \hline
%%a & b  & c  & d  & e  \\ \hline
%   & Beijing Taxi & San Francisco Cabs & Tokyo Flow & Chukyo Flow \\
%\hline
%STAY       & 0.378  & 0.346 & 0.186 & 0.187    \\
%CGM       & 0.301  & 0.307  & 0.181 & 0.448 \\
%CNP       & 0.375  & 0.296  & 0.182 & 0.179        \\
%SBP-EM    & 0.344  & 0.291 & 0.347  & 0.375  \\
%\hline
%SBP-ISTC   & $\mathbf{0.244}$  & $\mathbf{0.167}$ & 0.166 & 0.145   \\
%SBP-ISTA   & 0.253  & 0.187 & $\mathbf{0.156}$   & $\mathbf{0.129}$ \\
%\hline
% \end{tabular}
%\end{table*}

%\begin{table*}[htbp]
%\small
%\centering
%\begin{tabular}{c|c|c|c|c}
%\hline
%   & Beijing Taxi & San Francisco Cabs & Tokyo Flow & Chukyo Flow \\
%%\hline
%%STAY       & 0.378  & 0.346 & 0.186 & 0.187    \\
%%CGM       & 0.301  & 0.307  & 0.181 & 0.448 \\
%%CNP       & 0.375  & 0.296  & 0.182 & 0.179        \\
%%SBP-EM    & 0.344  & 0.291 & 0.347  & 0.375  \\
%%\hline
%%SBP-ISTC   & $\mathbf{0.244}$  & $\mathbf{0.167}$ & 0.166 & 0.145   \\
%%SBP-ISTA   & 0.253  & 0.187 & $\mathbf{0.156}$   & $\mathbf{0.129}$ \\
%\hline
%\end{tabular}
%\caption{Normalized mean absolute error (NAE) for the estimation of transition flows in the real-world datasets}
%\label{realdata}
%\end{table*}

\section{Experiments}

\subsection{Synthetic data}
% We conducted  to demonstrate the performance of the developed approximate EM algorithm in learning and estimating ensemble flows, compared with related state-of-the-art methods. We also compared the running time of the developed SBP-GS-EM algorithm with SBP-EM algorithm. Some additional experiments are implemented to demonstrate the ability of SBP enhanced graph neural networks in correcting SBP algorithm when sufficient amount of training samples are available.
%
% \section{Simulated data
\begin{figure*}[t]
  \centering
      \includegraphics[width=18cm,height=6.5cm, keepaspectratio]{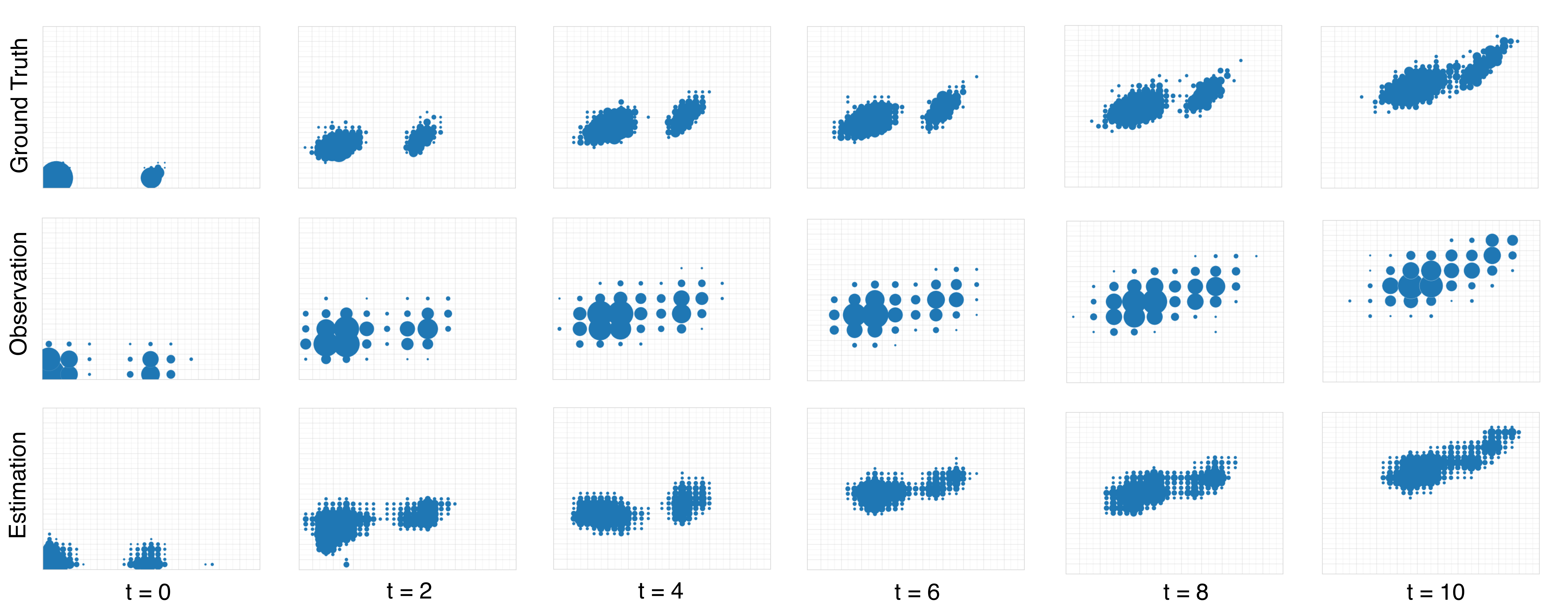}
      %\includegraphics[width=11cm,height=15cm, keepaspectratio]{figure/IRQ-IRN.pdf}
      %\includegraphics[width=14cm,height=15cm, keepaspectratio]{figure/ISR-LEB.pdf}
      %\vspace{-0.51em}
\caption{The top plots shows a simulated ensemble flow of $1,000$ particles moving over a $30\times30$ grid cells over $10$ time points; the middle displays the noisy observations of this ensemble flow; the bottom shows the distributions estimated by the proposed method for each corresponding time point. The size of the blue dots is proportional to the number of particles at the corresponding state.}
\label{flow}
\end{figure*}
\begin{figure}[ht]
  \centering
      \includegraphics[width=4.5cm,height=6cm, keepaspectratio]{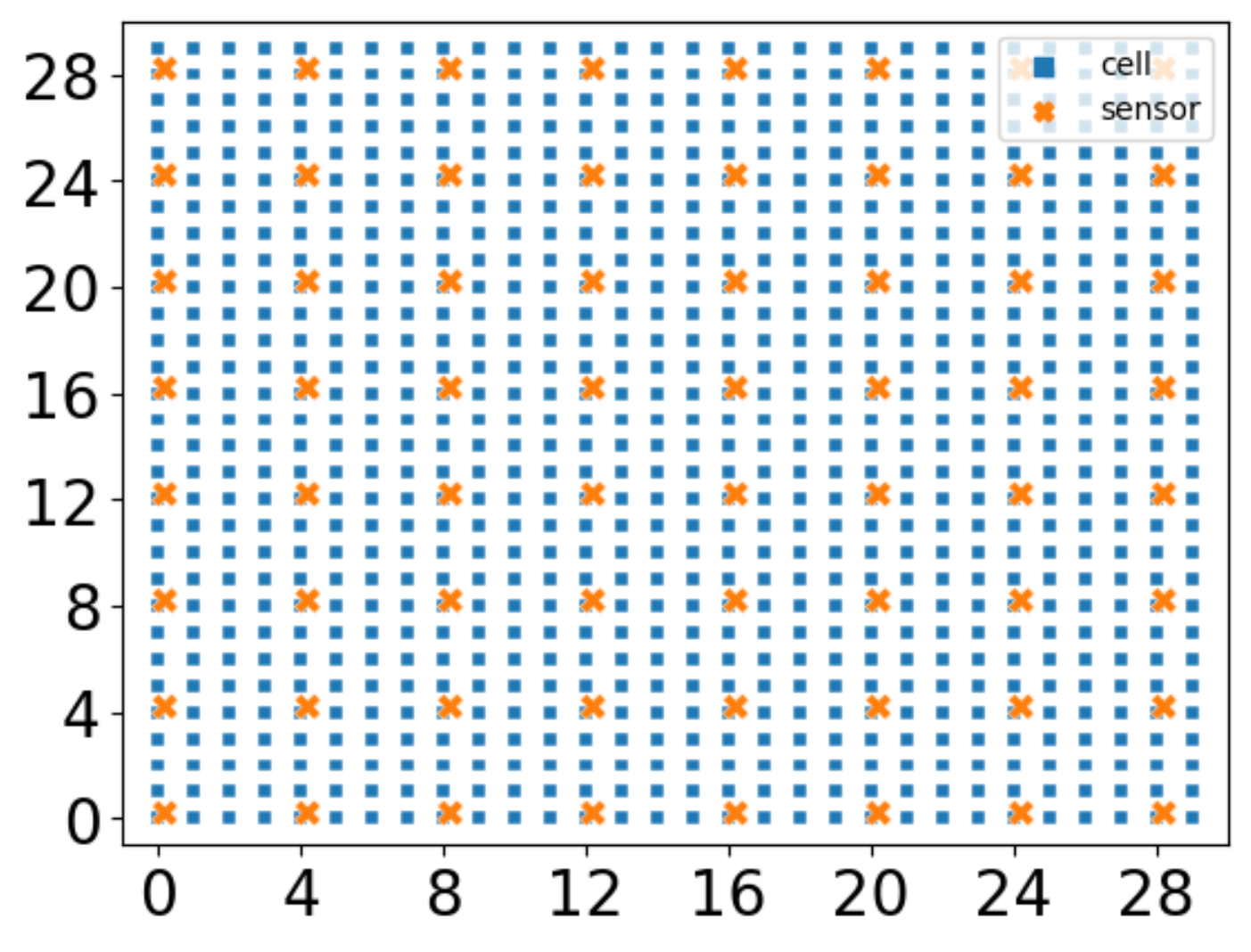}
      %\vspace{-0.51em}
\caption{Sensor locations}%(a) shows a simulated ensemble flows of $1000$ particles moving over a $12\times12$ grid over $15$ time points; (b) displays the noisy observations of this ensemble flow; (c) shows the particle distributions estimated by the SBP-GS-EM algorithm for each corresponding time point. The size of the blue dots is proportional to the number of particles at the corresponding state.}
\label{sensorlocations}
\end{figure}

Following the simulation studies~\cite{HaaslerRCK19,MOTSBP,MOTPGM}, we consider simulating an ensemble of $M$ individuals moving over a $30\times 30$ grid cells, as shown in Fig.~\ref{flow}. The goal of these individuals is to move from bottom-left and bottom middle corners to top-right corner. In particular, the dynamic behaviors of these individuals are determined by a log-linear distribution, which is modeled by four factors: the physical distance between two states, the angle between the moving direction and an external force, the angle between moving direction and the direction to the destination, and the preference to stay in the original state. The parameters of the log-linear model for these four factors, are set to be $(3,5,5,10)$, respectively. 
% In general, the collective dynamics of ensemble particles can be influenced by various external forces, which cannot be easily estimated. 
% In real population flow analysis, most people flow data are indirectly measured using various sensing devices. For instance, Wi-Fi hotspots, and cell phone based stations, can be roughly used to measure the number of individuals connected to them. Similarly, in our case, t
There are 64 sensors placed over the grids as shown in Fig.~\ref{sensorlocations}. 
Instead of collecting the full trajectories of all the particles, these sensors can only measure an aggregated count of individuals currently being observed. The probability of an individual being detected, decreases exponentially as the distance between the individual and the sensor increases.

 As shown in Fig.~\ref{flow}, although the sensor observations only roughly record the aggregated counts of individuals, the proposed method still well estimate the population flows with a high resolution.

\begin{table*}[htbp]
 \centering
 %\caption{Test table.}
 \caption{Normalized mean absolute error (NAE) for the estimation of transition flows in the real-world datasets}
 \label{realdata}
 %\small
 \begin{tabular}{ c|c|c|c|c } \hline
%  & t1 & t2 & t3 & t4 \\ \hline
%a & b  & c  & d  & e  \\ \hline
   & Beijing Taxi & San Francisco Cabs & Tokyo Flow & Chukyo Flow \\
\hline
STAY       & 0.378  & 0.346 & 0.186 & 0.187    \\
CGM       & 0.301  & 0.307  & 0.181 & 0.448 \\
CNP       & 0.375  & 0.296  & 0.182 & 0.179        \\
SBP-EM    & 0.344  & 0.291 & 0.347  & 0.375  \\
\hline
SBP-ISTC   & $\mathbf{0.244}$  & $\mathbf{0.167}$ & 0.166 & 0.145   \\
SBP-ISTA   & 0.253  & 0.187 & $\mathbf{0.156}$   & $\mathbf{0.129}$ \\
\hline
 \end{tabular}
\end{table*}

\subsection{Real-world data}
The performance of the proposed methods in estimating transition flows from aggregated count data, is evaluated using four real-world population flow data. The Beijing Taxi data~\cite{TDrive} consists of $10,357$ taxi trajectories collected from February 2, 2008 to February 8, 2008. The grid sizes in this data are 2km$\times$2km(17$\times$17 grid cells), and thus the number of states is 289. The time grid is 15 minutes, and thus the aggregated observations were made for 96 time steps for one day. The second data collected 537 taxi cabs' GPS traces in San Francisco from to May 18 2008 to May 30 2008. The grid sizes in this data are 2km$\times$2km(13$\times$13 grid cells), and thus the number of states is 169. The time grid is 15 minutes, and thus the aggregated observations were made for 96 time steps for one day. The Tokyo People Flow data\footnote{Data sources: SNS-based People Flow Data, http:
//nightley.jp/archives/1954} consists of 6,432, 9,166, 6,822, 10,134, 6,646, 10,338 individual trajectories on six days in the year of 2013. The grid sizes in this data are 10km$\times$10km(15$\times$15 grid cells), and thus the number of states is 225. The time grid is 30 minutes, and thus the aggregated observations were made for 48 time steps for one day. The Chukyo Flow data cosists of 975, 1,372, 1,195, 1,506, 1,021, 1,615 individuals also on six days in the year of 2013. The data is created in the same ways as Tokyo Flow data, except the grid sizes are 10km$\times$10km($10\times10$) grid cells.

The proposed methods were evaluated in terms of estimating the transition flows $\{\mathbf{M}^t\}_{t=1}^{T-1}$ only using the marginally aggregated count observations $\{\bm{\mu}_t\}_{t=1}^{T}$. The performance in estimating transition flows is evaluated using the normalized mean absolute error (NMAE) defined by
\begin{align}
\mathrm{NMAE} = \frac{\sum_{t=1}^{T-1}\sum_{i=1}^S\sum_{j\in\mathcal{N}_i}|\hat{M}^t_{ij}-\bar{M}^t_{ij}|}{\sum_{t=1}^{T-1}\sum_{i=1}^S\sum_{j\in\mathcal{N}_i} \bar{M}^t_{ij}},\nonumber
\end{align}
where $\mathcal{N}_i$ stands for the set of neighbor states of state $i$, $\bar{M}^t_{ij}$ is the true number of transitions from state $i$ at time $t$ to state $j$ at time $t+1$, and $\hat{M}^t_{ij}$ denotes the corresponding estimate.

\noindent\textbf{Baselines.}
The proposed methods were compared with some closely related methods: the collective graphical model~\cite{Bernstein16}, constrained norm-product (CNP) algorithm~\cite{MOTPGM}, and Sinkhorn belief propagation-Expectation Maximization (SBP-EM) algorithm~\cite{Chen02}. The STAY method assumes that all the individuals stay in the same states from time $t$ to time $t+1$, i.e., $\hat{M}^t_{ii} = (\bm{\mu}_t)_i$ and $\hat{M}^t_{ij} = 0$ for $j\neq i$. Both the CGM and SBP-EM algorithm assume the underlying Markov chains are time-homogeneous, while our proposed methods can estimate time-varying transition probabilities by learning the underlying cost matrices. 

\noindent\textbf{Results.} The normalized absolute error averaged over all the time steps for each data, is presented in Table~\ref{realdata}. For all the datasets, the proposed methods achieved higher accuracy than the other methods. In particular, we found that our proposed methods outperform the closely related SBP-EM algorithm by allowing the underlying cost matrices to be time-varying. In addition, we found that the SBP-ISTA performed better than SBP-ISTC in estimating transition flows in the Tokyo and Chukyo People Flow data. We looked into this data, and found that most individuals were moving from outer suburb regions to inner downtown areas in the morning, while transitioning on the opposite direction in the evening. The transition flows collected in this data, exhibit asymmetric moving patterns at different time steps. Hence, SBP-ISTA achieved higher accuracy by constructing more complicated cost matrices, compared with SBP-ISTC that enforces symmetric structured cost matrices.

\section{Conclusion}
This paper proposed to estimate population transition flows from marginally aggregated data via a graphical-structured multi-marginal optimal transport framework. More specifically, the proposed methods allow the transition kernels behind population flows to be time-varying, by learning the time-dependent cost functions. The uniqueness of the solutions is guaranteed under mild conditions. The experiments on four real-world population flow data, show the improved accuracy of the proposed methods in estimating latent transition flows, compared with the others built upon time-homogeneous Markov chains.

\subsection*{Acknowledgements}
We thank Xiaojing Ye and the anonymous reviewers for the many useful comments that improved this manuscript. 

\subsection*{Appendix: Inverse Multi-marginal optimal transport} 
Given the estimated transport plan $\mathbf{\hat{M}}$, we consider learning the transition parameters $\mathbf{A}^t$ by learning the corresponding cost matrices $\mathbf{\mathbf{C}}^t$, which can be effectively resolved via the inverse multi-marginal optimal transport (IMOT) formulation.
The IMOT problem can be written as
\begin{align*}
    \min_{\mathbf{C}}\quad &\  H(\hat{\mathbf{M}} | \mathbf{M}^*) + R(\mathbf{C}), \\
    \mbox{s.t.}\quad&\  \mathbf{M}^*\equiv \argmin_{\mathbf{M}\in\Pi(\mu_1,\dots,\mu_J)} \langle \mathbf{C}, \mathbf{M} \rangle + \epsilon H(\mathbf{M}\mid \mathbf{1}_{S\times\cdots\times S}),%\langle \mathbf{M}, \log \mathbf{M}
\end{align*}
where %$\hat{\mathbf{M}}$ denotes the estimated plan, and 
$R(\mathbf{C})$ is the regularization on $\mathbf{C}$. 
Let $\bm{\alpha}_j$ be the multiplier of the $j$th marginal equality constraint for $j=1,\dots,J$, and write the dual problem of lower-level problem $\mathbf{M}^*$ with given $\mathbf{C}$ as 
\begin{equation*}
    \max_{\bm{\alpha}_j}\ \sum_{j=1}^{J} \langle \bm{\alpha}_j, \bm{\mu}_j \rangle - \epsilon \langle \mathbf{K}, \mathbf{B}\rangle.
\end{equation*}
where $\mathbf{K}\equiv\exp(-\frac{\mathbf{C}}{\epsilon})$, $\mathbf{B}\equiv\mathbf{b}_1\otimes\cdots\otimes\mathbf{b}_T$ where $\mathbf{b}_t\equiv\exp(\frac{\bm{\alpha}_t}{\epsilon})$. Then, the corresponding optimal solution of the primal problem is
\begin{equation*}
    \mathbf{M}^* = \mathbf{K}\odot \mathbf{B}%e^{\sum_{j}\bm{\alpha}_{j}^* - \mathbf{C}} \in \mathbb{R}^{p_1\times \dots \times p_J},
\end{equation*}
which should be interpreted as 
\begin{equation*}
    (\mathbf{M}^*)_{i_1,\dots,i_J} = e^{\sum_{j=1}^J\bm{\alpha}_{j}^*(i_j) - \mathbf{C}(i_1,\dots,i_J)},
\end{equation*}
and $\bm{\alpha}_j^*(i)$ is the $i$th component of $\bm{\alpha}_j^*$.
Plugging this into the upper-level problem of IMOT, we have
\begin{equation*}
    \min_\mathbf{C}\ \langle \hat{\mathbf{M}}, \mathbf{C} \rangle - \sum_{j} \langle \bm{\alpha}_j^*, \bm{\mu}_j \rangle + R(\mathbf{C}).
\end{equation*}
Recalling the optimality of $\bm{\alpha}_j^*$:
\begin{equation*}
    -\langle \bm{\alpha}_j^*, \bm{\mu}_j \rangle + \epsilon = \min_{\bm{\alpha}_j}\ -\sum_{j=1}^{J} \langle \bm{\alpha}_j, \bm{\mu}_j \rangle + \epsilon \langle \mathbf{K}, \mathbf{B}\rangle ,
\end{equation*}
we obtain the unconstrained minimization problem equivalent to IMOT as
\begin{equation*}
    \min_{\mathbf{C},\bm{\alpha}_j}\ \langle \hat{\mathbf{M}}, \mathbf{C} \rangle - \sum_{j} \langle \bm{\alpha}_j, \bm{\mu}_j) + \epsilon \langle \mathbf{K}, \mathbf{B}\rangle + R(\mathbf{C}).
\end{equation*}

%\small
\bibliographystyle{named}
\bibliography{ref}
\end{document}